\documentclass{ecai} 

\usepackage{latexsym}
\usepackage{amssymb}
\usepackage{amsmath}
\usepackage{amsthm}
\usepackage{booktabs}
\usepackage{enumitem}
\usepackage{graphicx}
\usepackage{color}

\usepackage[export]{adjustbox}
\usepackage{xcolor}
\usepackage{cleveref}
\usepackage{xspace}
\usepackage{natbib}
\usepackage{accents}
\usepackage{algorithm}
\usepackage[noend]{algpseudocode}

\usepackage{silence}

\DeclareFontShape{T1}{ptm}{m}{scit}{<-> ssub * ptm/m/sc}{}

\newcommand\thickbar[1]{\accentset{\rule{.6em}{.1pt}}{#1}}

\newcommand{\Cstar}{C\ensuremath{^*}}
\newcommand{\CStar}{C\ensuremath{^*}}
\newcommand{\open}{\textsc{Open}\xspace}
\newcommand{\openF}{\textsc{Open\textsubscript{F}}\xspace}      % Open_F
\newcommand{\openf}{\textsc{Open\textsubscript{F}}\xspace}      % Open_F
\newcommand{\openB}{\textsc{Open\textsubscript{B}}\xspace}      % Open_B
\newcommand{\openb}{\textsc{Open\textsubscript{B}}\xspace}      % Open_B
\newcommand{\openD}{\textsc{Open\textsubscript{D}}\xspace}      % Open_D
\newcommand{\opend}{\textsc{Open\textsubscript{D}}\xspace}      % Open_D
\newcommand{\openo}{\textsc{Open\textsubscript{$\thickbar{D}$}}\xspace}      % Open_O

\newcommand{\closed}{\textsc{Closed}\xspace}
\newcommand{\closedf}{\textsc{Closed\textsubscript{F}}\xspace}      % Closed_F
\newcommand{\closedb}{\textsc{Closed\textsubscript{B}}\xspace}      % Closed_B
\newcommand{\closedD}{\textsc{Closed\textsubscript{D}}\xspace}      % Closed_D
\newcommand{\closedd}{\textsc{Closed\textsubscript{D}}\xspace}      % Closed_D
\newcommand{\start}{\ensuremath{start}\xspace}
\newcommand{\goal}{\ensuremath{goal}\xspace}

\newcommand{\BiHS}{\mbox{BiHS}\xspace}     % Bidirectional Heuristic Search
\newcommand{\UniHS}{\mbox{UniHS}\xspace}   % Unidirectional Heuristic Search
\newcommand{\pembfs}{\mbox{PEM-BiHS}\xspace}
\newcommand{\PEMBFS}{\mbox{PEM-BiHS}\xspace}

\newcommand{\mm}{\mbox{MM}\xspace}
\newcommand{\MM}{\mbox{MM}\xspace}
\newcommand{\bae}{\mbox{BAE$^*$}\xspace}
\newcommand{\BAE}{\mbox{BAE$^*$}\xspace}

\newcommand{\IDAStar}{\mbox{{IDA$^*$}}\xspace}
\newcommand{\AIDAStar}{\mbox{{AIDA$^*$}}\xspace}
\newcommand{\AIDARStar}{\mbox{{rAIDA$^*$}}\xspace}
\newcommand{\astar}{\mbox{{A$^*$}}\xspace}

\newcommand{\pemm}{\mbox{PEMM}\xspace}
\newcommand{\PEMM}{\mbox{PEMM}\xspace}
\newcommand{\PEMBAE}{\mbox{PEM-BAE$^*$}\xspace}
\newcommand{\PEMASTAR}{\mbox{PEM-A$^*$}\xspace}
\newcommand{\RPEMSTAR}{\mbox{PEM-rA$^*$}\xspace}
\newcommand{\PIDA}{\mbox{AIDA$^*$}\xspace}
\newcommand{\RPIDA}{\mbox{rAIDA$^*$}\xspace}

\newcommand{\gD}{\mbox{$g_D$}}
\newcommand{\gF}{\mbox{$g_F$}}
\newcommand{\gB}{\mbox{$g_B$}}

\newcommand{\gminf}{\mbox{$gMin_F$}\xspace}     % lowest g value in Open_F
\newcommand{\gminb}{\mbox{$gMin_B$}\xspace}     % lowest g value in Open_B

\newcommand{\hD}{\mbox{$h_D$}}
\newcommand{\hF}{\mbox{$h_F$}}
\newcommand{\hB}{\mbox{$h_B$}}

\newcommand{\fD}{\mbox{$f_D$}}
\newcommand{\fF}{\mbox{$f_F$}}
\newcommand{\fB}{\mbox{$f_B$}}

\newcommand{\fminf}{\mbox{$fMin_F$}\xspace}     % lowest f value in Open_F
\newcommand{\fminb}{\mbox{$fMin_B$}\xspace}     % lowest f value in Open_B
\newcommand{\prmin}{\mbox{$PrMin$}\xspace}

\newcommand{\dF}{\mbox{$d_F$}}
\newcommand{\dB}{\mbox{$d_B$}}

\newcommand{\bF}{\mbox{$b_F$}}
\newcommand{\bB}{\mbox{$b_B$}}

\newcommand{\bminf}{\mbox{$bMin_F$}\xspace}     % lowest b value in Open_F
\newcommand{\bminb}{\mbox{$bMin_B$}\xspace}     % lowest b value in Open_B
\newcommand{\bmind}{\mbox{$bMin_D$}\xspace}

\begin{document}

%%%%%%%%%%%%%%%%%%%%%%%%%%%%%%%%%%%%%%%%%%%%%%%%%%%%%%%%%%%%%%%%%%%%%%%%

\begin{frontmatter}

%%% Use this command to specify your submission number.
%%% In doubleblind mode, it will be printed on the first page.

\paperid{1331} 

%%% Use this command to specify the title of your paper.

\title{On Parallel External-Memory Bidirectional Search}

%%% Use this combinations of commands to specify all authors of your 
%%% paper. Use \fnms{} and \snm{} to indicate everyone's first names 
%%% and surname. This will help the publisher with indexing the 
%%% proceedings. Please use a reasonable approximation in case your 
%%% name does not neatly split into "first names" and "surname".
%%% Specifying your ORCID digital identifier is optional. 
%%% Use the \thanks{} command to indicate one or more corresponding 
%%% authors and their email address(es). If so desired, you can specify
%%% author contributions using the \footnote{} command.

\author[A]{\fnms{Lior}~\snm{Siag}}
\author[A]{\fnms{Shahaf S.}~\snm{Shperberg}\thanks{Email: shperbsh@bgu.ac.il}}
\author[A]{\fnms{Ariel}~\snm{Felner}}
\author[B,C]{\fnms{Nathan R.}~\snm{Sturtevant}} 

\address[A]{Ben-Gurion University of the Negev}
\address[B]{University of Alberta, Department of Computing Science}
\address[C]{Alberta Machine Intelligence Institute (Amii)}

%%% Use this environment to include an abstract of your paper.

\begin{abstract}
Parallelization and External Memory (PEM) techniques have significantly enhanced the capabilities of search algorithms when solving large-scale problems. 
Previous research on PEM has primarily centered on unidirectional algorithms, with only one publication on bidirectional PEM that focuses on the meet-in-the-middle (MM) algorithm.
Building upon this foundation, this paper presents a framework that integrates both uni- and bi-directional best-first search algorithms into this framework. We then develop a PEM variant of the state-of-the-art bidirectional heuristic search (\BiHS) algorithm BAE* (PEM-BAE*). As previous work on \BiHS did not focus on scaling problem sizes, this work enables us to evaluate bidirectional algorithms on hard problems. Empirical evaluation shows that PEM-BAE* outperforms the PEM variants of A* and the MM algorithm, as well as a parallel variant of IDA*. These findings mark a significant milestone, revealing that bidirectional search algorithms clearly outperform unidirectional search algorithms across several domains, even when equipped with state-of-the-art heuristics.
\end{abstract}

\end{frontmatter}

%%%%%%%%%%%%%%%%%%%%%%%%%%%%%%%%%%%%%%%%%%%%%%%%%%%%%%%%%%%%%%%%%%%%%%%%

\section{Introduction}

\astar~\citep{hart1968formal} and its many variants are commonly used to optimally solve combinatorial and pathfinding problems. However, as these problems often involve state spaces of exponential size, practical limitations in terms of time and memory resources hinder the ability of these algorithms to tackle large instances, particularly when the scale of the problem increases (e.g., more variables, larger graphs, etc.). Thus, an important line of research in heuristic search is harnessing hardware capabilities to overcome these limitations. Parallelizing various components of the search process across multiple threads can lead to a significant reduction in running time. Similarly, exploiting external memory, such as large disks, allows algorithms to scale the size of \open and \closed lists, substantially increasing the size of problems that can be solved \citep{dobbelin2014building,edelkamp2012distributed,edelkamp2004external,hu2019emreversed,kunkle2007twenty}.

Orthogonally, recent research has focused on {\em bidirectional heuristic search} (\BiHS) algorithms, which have been shown to outperform unidirectional search (\UniHS) methods \citep{alcazar20unifying, front2017chen, shperberg2019enriching, siag2023front}, on some problems, but the majority of these results are on relatively small problems or with weakened heuristics. It has been conjectured \cite{barker2015limitations} that bidirectional search does not perform well with strong heuristics, and it is unclear whether these results will scale to the largest problems. Thus, our aim is to scale bidirectional search algorithms to significantly larger problems and stronger heuristics.

The intersection of parallel and external memory (PEM) and \BiHS has only been explored in the context of the meet-in-the-middle (\MM) algorithm~\citep{holte2017mm}, yielding a variant called \pemm~\citep{external2016sturtevant}. However, utilizing recent advancements in \BiHS algorithms necessitates a framework for seamlessly converting both recent and future \BiHS algorithms into corresponding PEM variants. 

This paper builds upon previous research in parallel and external-memory search~\citep{external2016sturtevant} to unify and explore how parallel computing and external memory utilization impact \BiHS. In the interest of broad accessibility, we introduce a flexible framework capable of integrating any \UniHS or \BiHS algorithm into the PEM paradigm. Additionally, using this framework, we successfully integrate the state-of-the-art \BiHS algorithm \bae~\citep{sadhukhan2013bidirectional}, which relies on a consistent heuristic, resulting in a variant called \PEMBAE. 

We empirically compare \PEMBAE against PEM variants of \astar (\PEMASTAR), reverse A* (\RPEMSTAR), and \MM (\pemm) %\ns{the previously published algorithm can't be renamed})
on 15-and 24-tile puzzles and random 20-disk 4-Peg Towers of Hanoi problems. In addition, we compare \PEMBAE to a parallel variant of \IDAStar, called \AIDAStar~\citep{reinefeld1994work}, as well as its reversed version. 
Our results confirm that \PEMBAE significantly outperforms the other algorithms as problem complexity increases. This superiority of \PEMBAE is evident not only with weak heuristics, as previously demonstrated for \BAE, but also persists when utilizing strong heuristics on large problems. This highlights that a \BiHS~algorithm stands as the state-of-the-art approach for addressing multiple challenging problems.

\section{Background and Definitions}

Research on bidirectional heuristic search (\BiHS) has spanned several decades, dating back to the work of ~\citet{Pohl1971a}. Recently, a new theoretical understanding emerged regarding the necessary nodes for expansion during the search~\citep{sufficient2017eckerle,shaham2017minimal}, sparking a series of algorithms that find optimal~\citep{alcazar20unifying, barley2018gbfhs, front2017chen, shperberg2019enriching}, near-optimal~\citep{atzmon2023wrestrained}, and memory-efficient solutions~\citep{shperberg2021iterative}. Other research has delved into algorithm comparisons~\citep{alcazar20unifying,shperberg2020differences,siag2023front} and explored the potential advantages of \BiHS over Unidirectional search (\UniHS)~\citep{sturtevant2020predicting}. 

The focus of this paper is on two algorithms: \MM, the first algorithm to be integrated with parallel external-memory search, and \BAE, chosen due to its superior performance. \BAE and other \BiHS algorithms have demonstrated strong performance, surpassing their \UniHS counterparts, on small problems with relatively weak heuristics. Some have suggested that, as the scale of problems increase, \BiHS algorithms can maintain their dominance over \UniHS algorithms, even when stronger heuristics are employed \citep{alcazar20unifying,siag2023comparing,sturtevant2020predicting}. Others have conjectured that bidirectional search will not perform well on problems with strong heuristics \cite{barker2015limitations}.
This paper evaluates these conjectures by constructing a PEM variant of \BAE capable of solving large problems with state-of-the-art heuristics.

\subsection{\BiHS: Definitions and Algorithms}

In \BiHS, the aim is to find a least-cost path, of cost \CStar, between \start and \goal in a given graph $G$. $\mathit{dist}(x,y)$ denotes the shortest distance between $x$ and $y$, so $\mathit{dist}(start, goal)= \Cstar$. \BiHS executes a forward search (F) from \start and a backward search (B) from \goal until the two searches meet. \BiHS\ algorithms typically maintain two open lists \openf and \openb for the forward and backward searches, respectively. Each node is associated with a $g$-value, an $h$-value, and an $f$-value ($g_F, h_F, f_F$ and $g_B, h_B, f_B$ for the forward and backward searches). 
Given a direction $D$ (either F or B), we use \fD, \gD\ and \hD\ to indicate $f$-, $g$-, and $h$-values in direction $D$. 

The $g$-value of a state $s$ is cost of the best path discovered to $s$, and $f$-value of a state is the sum of its $g$- and $h$- values.
Most \BiHS algorithms consider the two {\em front-to-end} heuristic functions~\citep{Kaindl1997} $\hF(s)$ and $\hB(s)$ which respectively estimate $\mathit{dist}(s,goal)$ and $\mathit{dist}(\start,s)$ for all $s \in G$. $h_F$ is \emph{forward admissible}\ iff
$h_F(s)\le \mathit{dist}(s,goal)$ for all $s$ in $G$ and is \emph{forward consistent}\ iff
$h_F(s)\le \mathit{dist}(s,s^\prime)+h_F(s^\prime)$ for all $s$ and $s^\prime$ in~$G$.
Backward {\em admissibility} and {\em consistency} are defined analogously.

\BiHS algorithms mainly differ in their node- and direction-selection strategies and other termination criteria. We next describe \MM and \BAE, both implemented in this paper.

\paragraph{Meet in the middle (\mm).}
\mm~\citep{holte2017mm} is a \BiHS algorithm ensuring that the search frontiers {\em meet in the middle}. In \mm, nodes $n$ in \openD are prioritized by: 
\begin{equation} \label{eq:MM}
    pr_D(n)=\max(f_D(n),2g_D(n))
\end{equation}
\mm expands the node with minimal priority, $\prmin$ on both \openf and \openb. \mm halts the search once the following two conditions are met:  (1) the same node $n$ is found on both \open lists; (2) the cost of the path from \start to \goal through $n$ is $\leq LB_{\text{MM}}$ where $LB_{\text{MM}}$ is a lower bound on $\Cstar$ that is computed as follows:
\begin{equation} \label{eq:MM-U}
    LB_{\text{MM}} = \max(\prmin, \fminf, \fminb , \gminf + \gminb)
\end{equation}
where \fminf (\fminb) and \gminf (\gminb), are the minimal $f$- and $g$-values in \openf (\openB), respectively. \citet{external2016sturtevant} provided a PEM variant of \MM (\pemm). Our framework below is an extension of \pemm.

\paragraph{BAE*.}
Most algorithms (e.g., \mm) only assume that the heuristics used are admissible. \BAE~\citep{sadhukhan2013bidirectional,alcazar20unifying} (and the identical algorithm DIBBS~\citep{SewellJ21}) are \BiHS algorithms that specifically assume that both \hF\ and \hB\ are consistent and thus exploit this fact.  %\ss{do we need to mention DIBBS? AF: yes we do}
Let $\dF(n) = \gF(n) - \hB(n)$, the {\em difference} between the actual forward cost $n$ (from \start) and its heuristic estimation to \start.  This indicates the {\em heuristic error} for node $n$ (as $\hF(n)$ is a possibly inaccurate estimation of $\gB(n)$). Likewise,  $\dB(m) = \gB(m) - \hF(m)$. \BAE orders nodes in \openF  according to 
\begin{equation} \label{eq:BAE}
    \bF(n) = \fF(n)+\dF(n)
\end{equation}

$\bF(n)$ adds the heuristic error $\dF(n)$ to $\fF(n)$ to indicate that the opposite search using $\hB(n)$ will underestimate by $\dF(n)$.  Likewise, $\bB(m) = \fB(m) + \dB(m)$ is used to order nodes in \openB. At every expansion cycle, \BAE chooses a search direction $D$ and expands a node with minimal $b_D$-value. Additionally, \BAE terminates once the same state $n$ is found on both \open lists and the cost of the path from \start to \goal through $n$ is $\leq LB_B$ where $LB_B$, is a the following lower bound on $\Cstar$ (known as the $b$-bound):
\begin{equation} \label{eq:BAE-U}
    LB_{B} = {(\bminf + \bminb)}/{2}
\end{equation}
where \bmind is the minimal $b$-value in \opend.

Given a consistent heuristic, \bae was proven to return an optimal solution. $b(n)$ is more informed than other priority functions (as it also considers $d(n)$), and was shown to outperform common unidirectional and bidirectional algorithms~\citep{alcazar20unifying,siag2023front} on relatively simple domains. These included the 15-puzzle with the Manhattan Distance heuristic, the 12-disk 4-peg Towers of Hanoi Problem, grid benchmark problems, and the pancake puzzle with a weakened GAP heuristic.
While \bae stands as a state-of-the-art \BiHS algorithm, it has not been tested on large sliding-tile or Towers of Hanoi problems, or with large pattern database heuristics. In this paper, we develop \PEMBAE, a PEM version of \BAE, which is able to scale to far larger problems and heuristics.

\subsection{Parallel External-Memory Search}

External Memory Search structures a search such that the maximum size of a problem solved scales according to the size of the disk, instead of available RAM \citep{kunkle2008solving}. These algorithms are often paired with parallel search methods, as techniques to minimize random I/O often group states together, allowing parallel processing.

Two classes of external memory search appear in the literature. One class aims to perform a complete breadth-first search of a state space, employed for verifying state space properties \citep{korf2008linear} or building large heuristics \citep{hu2019emreversed}. These approaches maintain information about every state on disk, loading portions of the data into memory for expansion and duplicate detection. 
Another class of algorithms, including External A* \citep{edelkamp2004external} and search with structured duplicate detection \citep{zhou2004structured,zhou2007parallel}, is used to solve large problem instances. In these algorithms, \open is stored explicitly on disk, and \closed may or may not be stored, depending on the properties of the state space \citep{korf2005frontier}.

Both classes aim to reduce I/O operations to disk, e.g., by delaying operations like duplicate detection until many states can be processed in parallel~\citep{best2004korf}, and dividing the state space up into smaller buckets of states~\citep{edelkamp2004external,hatem2018solving,korf2008minimizing,sturtevant2013minimizing, torralba2017efficient} which are stored together on disk, and then loaded and expanded together. When duplicate states all hash into the same bucket, it reduces the complexity of checking for duplicates. Hash-based duplicate detection~\citep{korf2008linear} and sorting-based duplicate detection~\citep{korf2016comparing} take buckets of states where duplicate detection has been delayed, load them into memory, and remove duplicates using hash tables or sorting, respectively. Structured duplicate detection~\citep{zhou2004structured} structures the state space so that all successor buckets can be loaded into memory for immediate duplicate detection.

External memory search has often relied on exponential-growing state-spaces with unit costs to ensure sufficient states are available to efficiently process in parallel. Notably, algorithms like PEDAL \citep{hatem2018solving} have extended these approaches to non-unit-cost problems.

\section{The \pembfs Framework}

We next introduce a high-level framework called {\em Parallel External Memory Bidirectional Heuristic Search} (\pembfs), into which both \BiHS and \UniHS can be seamlessly integrated. \pembfs is
designed to efficiently solve very large problem instances. Leveraging parallelization capabilities along with using external memory, \pembfs utilizes the foundations laid by algorithms such as \pemm, PEDAL, External A*, and structured duplicate detection and exploits further parallelization opportunities.

We begin with a high-level description of \PEMBFS together with the pseudo-code presented in \Cref{alg:pembfs}, which builds on \cite{external2016sturtevant}, followed by a detailed description of algorithmic components. \PEMBFS initializes an \open and \closed list for each direction (line~\ref{line:pembfs-initialize_open}). These lists do not explicitly store search nodes; instead, they maintain references to files (buckets) that contain the corresponding nodes. \PEMBFS employs the following stages:

\noindent {\bf Halting condition} (line~\ref{line:pembfs-termination}): During each expansion cycle, \PEMBFS evaluates the cost $U$ of the current incumbent solution in comparison to the calculated lower bound $LB$, derived from the nodes within the open lists. If $U \leq LB$ or one of the open lists is empty, \PEMBFS halts and returns the current solution cost (or infinity if no solution was found). Otherwise, the search continues. 

\smallskip
\noindent {\bf Choosing direction and bucket for expansion} (line~\ref{line:pembfs-get_direction}--\ref{line:pembfs-get_next_bucket}): next, \PEMBFS chooses the search direction $D$ (\UniHS always chooses the forward side)
as well as a bucket from \openD to be expanded.

\smallskip
\noindent {\bf Retrieving the bucket} (line~\ref{line:pembfs-read_bucket}):  Next \PEMBFS  performs parallel reading of the file containing the bucket from external memory into the internal memory (RAM). This stage involves eliminating duplicate states within the bucket.

\smallskip
\noindent {\bf Duplicate Detection (DD)} (line~\ref{line:pembfs-remove_duplicates}): \PEMBFS then eliminates duplicates nodes with  other buckets in \closedD.

\smallskip
\noindent {\bf Solution Detection} (line~\ref{line:pembfs-solution_detection}): \PEMBFS checks whether a new solution was found.

\smallskip
\noindent {\bf Expansion} (line~\ref{line:pembfs-expand_bucket} detailed in Algorithm~\ref{alg:expand_bucket}): Nodes from memory are concurrently expanded, generating children. These children are then written to their respective buckets (creating new ones if needed, and their references are inserted into \openD).

\smallskip
\noindent {\bf Writing to disk} (line~\ref{line:pembfs-write_closed}): Finally, the expanded nodes are written to disk, creating a new duplicate-free bucket. A reference to this bucket is inserted into \closed.

We next turn to provide an in-depth description of different algorithmic components of \PEMBFS.

\begin{algorithm}[t]
    \caption{PEM-BiHS General Framework}
    \label{alg:pembfs}
    \begin{algorithmic}[1] %[1] enables line numbers
        \Procedure{PEM-BiHS }{$start,goal$}
        \State $U\leftarrow\infty, LB\leftarrow $ ComputeLowerBound() \label{line:pembfs-initialize_cu}
        \State \openf, \openb, \closedf,\closedb $\leftarrow \emptyset$\label{line:pembfs-initialize_open}
        \State Push(\start, \openf) \Comment{create bucket and record}
        \State Push(\goal, \openb)
        \While {$\openF \neq \emptyset \wedge \openB \neq \emptyset \wedge U > LB$} \label{line:pembfs-termination}
            \State $D$ $\leftarrow$ ChooseDirection() \label{line:pembfs-get_direction}
            \State $b$ $\leftarrow$ ChooseNextBucket(\opend) \label{line:pembfs-get_next_bucket}
            \State ParallelReadBucket($b$, $D$) \Comment{including In-Bucket DD} \label{line:pembfs-read_bucket}
            \State RemoveDuplicates($b$, \closedd) \label{line:pembfs-remove_duplicates}
            \State CheckForSolution($U$, $b$, \openo) \label{line:pembfs-solution_detection} 
            \State ParallelExpandBucket($b$, \opend) \label{line:pembfs-expand_bucket}
            \State WriteToClosed($b$, \closedd)  \label{line:pembfs-write_closed}
            \State $LB \leftarrow$ ComputeLowerBound() \label{line:pembfs-lower_bound}
        \EndWhile
        \State \textbf{return} $U$
        \EndProcedure
    \end{algorithmic}
\end{algorithm}

\subsection{State-space Representation}
\label{ssec:state_rep_and_hash}

Typically, states are represented using high-level structures for convenient programming. In the context of combinatorial puzzles (e.g., the 15- and 24-puzzles), states are commonly stored as an array, with each cell's value corresponding to the label of the object it holds. To pack these states into files, an additional encoding/decoding mechanism is required for converting states into bits and decoding them. These decodings aim to minimize memory consumption and reduce I/O time.
For example, in the 24-puzzle practical implementations often use 8 bits (byte) to represent the identity of each tile demanding 200 bits ($8 \times 25$). Leveraging bit manipulation allows compression to 125 bits (5 bits are enough to store the identity of a tile). Furthermore, recognizing that a state is a permutation of 0-24 enables to use only $\left \lceil{\log_2(25!)}\right \rceil = 84$ bits, e.g., using the Lehmer encoding, which maps each permutation to a unique integer in the range $\{1 \ldots 25!\}$.

\subsection{Bucket Structure}

Nodes are grouped into buckets based on pre-determined attributes or \emph{identifiers}. E.g., in the context of External A*, a bucket groups nodes with identical $g$- and $h$-values (e.g., all nodes $n$ with $g(n) = 3$ and $h(n)=4$ belong to the $3$-$4$-bucket). In \PEMM, a bucket groups nodes with identical priority ($pr(n)$, as defined for \mm) as well as identical $g$-value. Alternatively, a hash value of a state, obtained by applying a hashing function to divide the set of states into a fixed number of values, can also serve as a property for defining a bucket. 

Importantly, \PEMBFS assumes that an entire bucket can fit into memory (in addition to a fixed-sized cache used to store successors before flushing them to disk, as detailed in Section~\ref{ssec:node_expansion}). Therefore, bucket identifiers may not allow too many nodes to be mapped into a single bucket, although adaptive methods have been used to dynamically adapt bucket sizes \citep{zhou2011dynamic}. Buckets are then written to files and loaded into memory as needed.

In \PEMBFS, \open and \closed are maintained inside main memory. They store bucket \emph{records}, which include the bucket identifiers and a link to the file containing the bucket. During an expansion cycle, a bucket record from \open is chosen for expansion. That bucket's file is then loaded into memory. Different algorithms within \PEMBFS will choose different buckets as described next.

\subsection{Direction, Prioritization, and Lower-bound} \label{ssec:dsbplb}

\textbf{Selecting direction} (line~\ref{line:pembfs-get_direction} in \Cref{alg:pembfs}). Numerous strategies can be employed for direction selection in \BiHS. Three prevalent strategies appear in the \BiHS literature: i) choosing the direction with minimal priority (as employed by \MM), ii) alternating between search directions, and iii) Pohl's cardinality criterion, which chooses the direction with the smaller open-list.  Naturally, \UniHS algorithms consistently choose the same direction.

\textbf{Prioritizing buckets} (line~\ref{line:pembfs-get_next_bucket}).
Selecting the next node to expand is the essence of any search algorithm. \pembfs allows the use of any priority function, under the restriction that the priority function must induce a total order among buckets based on the bucket identifiers. Thus, a bucket is chosen by comparing bucket identifiers within \open.  For example, in the implementation of a PEM variant of \astar using this framework (denoted as \PEMASTAR), the bucket identifier includes the $g$- and $h$-values of nodes. 

It is well known that the priority function influences the number of nodes expanded. For example, \astar prioritizes nodes based on lower $f$-values, and often adds a second prioritization criterion (``tie-breaking'' between nodes that share the same $f$-value), which prefers nodes with higher $g$-value (and thus lower $h$-value). This ``higher-$g$-first'' tie-breaking typically reduces the number of expanded nodes compared to ``lower-$g$-first", particularly in domains with unit edge costs. 
Nevertheless, in standard \astar, different tie-breaking policies usually expand states at the same rate (nodes per second). By contrast, in the context of \pembfs, tie-breaking policies can also significantly affect the node expansion rate. If PEM-A* breaks ties according to lower-$g$-first, once a $g$-$h$-bucket is expanded, new nodes will never be added to it, due to the monotonically increasing nature of $g$-values. By contrast, when using a higher-$g$-first policy, nodes can be added to the bucket after its expansion. Consequently, the same bucket can be selected for expansion multiple times, up to a number that scales quadratically with the total number of buckets, instead of only once. %[[AF: quadratic not clear]]
Thus, while using the lower-$g$-first tie-breaking may result in more expansions when compared to higher-$g$-first, the gains from minimizing the costly I/O operations by the lower-$g$-first tie-breaking make it a more resource-effective strategy. We have validated this in an empirical study on the 15-puzzle problem with a PDB heuristic (see details in Section~\ref{sec:experiments}). \PEMASTAR with the lower-$g$-first policy expanded 4.4 times more nodes than the higher-$g$-first policy (616,758 vs.
 2,724,974), but still was able to run four times faster (7.8 vs 32 seconds). 

\textbf{Computing Lower-bound} (line~\ref{line:pembfs-lower_bound}). Different lower-bounds may be employed by  \pembfs algorithms~\citep{siag2023front}. For instance, \astar uses the minimal $f$-value in the open list as a lower bound on the solution cost, \mm uses $LB_{\text{MM}}$ (Eq.~\ref{eq:MM-U}) and \bae uses $LB_B$ (Eq.~\ref{eq:BAE-U}). 
These and other bounds can be plugged into our framework. 

\subsection{Reading Buckets}
In prior PEM search studies, the process of reading a bucket (file) was carried out sequentially, influenced by the constraints of Hard Disk Drives (HDDs). Concurrent threads accessing the same file on HDDs could lead to performance degradation.
However, the recent, popular Solid-State Drives (SSDs) not only provide faster memory access but also benefit from parallelized reading. Therefore, \pembfs further optimizes performance by parallelizing the reading process. 
To parallelize the reading of a bucket, \pembfs distributes the file containing the bucket equally among multiple threads. In our experiments, this resulted in a twofold speed-up compared to sequential reading.
Subsequently, each thread reads states from the disk and decodes them into their in-memory state representation (See Section~\ref{ssec:state_rep_and_hash}). 

\subsection{In-bucket Duplicate Detection}

To minimize I/O operations, the elimination of duplicates is delayed until a bucket is chosen for expansion. Duplicate nodes may arise within the same bucket when a state is discovered via different paths. Additionally, duplicates can occur within closed buckets if the state has already been expanded with a lower $g$-value, or within other open buckets if states have been discovered with the same $g$-value. Notably, if the bucket identifiers include the $g$-value, other open buckets cannot contain duplicates of the same state with the same $g$-value, and can be disregarded for duplicate detection. During the reading phase, we manage in-bucket duplicate detection. However, to minimize I/O, we postpone duplicate detection within the closed list (as discussed in Section~\ref{ssec:duplicate_detection}) until after all nodes have been read.

Similar to hash-based delayed duplicate detection (DDD) ~\citep{best2004korf,korf2008linear}, when loading a bucket into memory, its nodes are placed into a hash table based on their states using a perfect hash function, and duplicates are ignored if their cell is already filled. 
Unlike the work of \citet{korf2008linear}, \PEMBFS allows multiple threads to read from the same bucket concurrently. To support this, each unique hash value is paired with a mutex. Once the reading and in-bucket duplicate detection is done, all threads are synchronized. 

\subsection{Duplicate Detection against \closed} \label{ssec:duplicate_detection}
To identify and eliminate duplicates of in-memory nodes with respect to nodes stored in closed buckets~(line~\ref{line:pembfs-remove_duplicates}), a scanning process is initiated which reads relevant buckets (in small increments) and compares them against the in-memory nodes. With {\em Delayed Duplicate Detection}~\cite{best2004korf}, each bucket is associated with a hash value and exclusively contains states assigned that particular value by a hash function. As a result, the scanning process is confined to closed buckets sharing the same hash value as the state of the node under consideration. For instance, if a node's state has a hash value of $5$, only buckets associated with the hash value $5$ are relevant and considered for duplicate detection.

In \pembfs, the determination of the relevant buckets for scanning relies on the identifiers that define them. If the bucket records include hash values of states, a similar approach to \citet{best2004korf} can be employed. Alternatively, if the bucket record contains the \hD-value of states, for direction $D$, as an identifier, only buckets with the same \hD-value need to be scanned. Therefore, if a bucket record contains both the \hF-value and the \hB-value, the scanning process is limited to buckets that possess identical values for both \hF\ and \hB. 

In addition, with unit edge-cost undirected graphs, (where edges can be followed in both ways) there are three cases for finding duplicates of a node generated at level $x$. (1) A parent $p$ at level $x-2$ generates a child node $n$ at level $x-1$. The child node $n$ generates its parent again at level $x$. (2) Consider a cycle of even length $k$ which was first explored by the search at the ancestor node $a$ at level 0. The farthest node of this cycle will be seen twice at level $x=k/2$ from two parents which are at level $x-1$. (3) Consider a cycle with odd length $k$. Here, the two nodes farthest from $a$ will be generated at level $y=\lfloor  k/2 \rfloor$ and each will generate the other at level $x=y+1$. Note that duplicates only occur as a result of a cycle, and a cycle can be either even or odd. Thus, when generating node $n$ with $g(n)=x$ we only need to check buckets with $g$-values of $x-2$, $x$, and $x-1$, for these three cases, respectively. 
Consequently, using the $g$-value as an identifier could significantly reduce the number of buckets that need to be scanned.
This approach optimizes the duplicate detection process based on the available information in the bucket records.

\subsection{Parallel Node Expansion}
\label{ssec:node_expansion}

\begin{algorithm}[tb]
    \caption{Parallel Bucket expansion pseudo-code}
    \label{alg:expand_bucket}
    \begin{algorithmic}[1] %[1] enables line numbers
    \Procedure{ParallelExpandBucket}{$b$, $D$}
        \State cache $\leftarrow \emptyset$ 
        \For{every state $s$ in $b$}
            \If{current thread should expand $s$}
                \For{each successor $s_i$ of $s$}
                    \State $sb$ $\leftarrow$ GetBucketOfNode($s_i$)
                    \If{$sb$ not in \opend}
                        \State AddBucket(\opend, $sb$)
                    \EndIf
                    \State AddState(cache$_{sb}$, $s_i$)
                    \If{cache\textsubscript{$sb$} is full}
                        \State{FlushToDisk(cache\textsubscript{$sb$})} \Comment{Also locks $sb$}
                    \EndIf
                \EndFor
                
            \EndIf
        \EndFor
        \If{cache is not empty}
            \State{FlushToDisk(cache)}
        \EndIf
    \EndProcedure
    \end{algorithmic}
\end{algorithm}

The process of parallel bucket expansion (line~\ref{line:pembfs-expand_bucket}) is outlined in \Cref{alg:expand_bucket}. To enable parallel expansion, each thread is allocated an equal portion of the nodes. Newly generated nodes are inserted into a dedicated successor cache for each thread.
Each cache is divided into smaller arrays, where each array corresponds to a specific bucket record.
This avoids the necessity of immediately writing each new successor to the file, thereby minimizing I/O operations.\footnote{One may perform a DD inside this cache array. This will have a marginal effect on the overall performance.}
The framework supports reopenings by adding a new bucket with the same identifier to \open and merging it with the corresponding \closed bucket after its expansion. In our evaluation, we used consistent heuristics, so we did not encounter any reopenings.
Once an array reaches its capacity, all its nodes are flushed into the disk, generating new buckets on disks as well as new bucket records in \open if necessary. To prevent simultaneous writes to the same file and maintain data integrity, each bucket record is linked to a mutex, which is locked when a thread writes to a bucket.
This design maintains data integrity and allows concurrent writing.

Note that bucket expansion and delayed solution detection in this framework (see \Cref{ssec:solution}) can occur concurrently with separate threads, as delayed solution detection is only concerned with the nodes that are about to be expanded, which are already loaded into memory and will not change due to the expansion. 

\subsection{Solution Detection} \label{ssec:solution}

There are two approaches for solution detection (Algorithm 1, line~\ref{line:pembfs-solution_detection}): {\em immediate solution detection} (ISD) and {\em delayed solution detection} (DSD).
In ISD, solutions are identified upon generation of a node $n$. This involves a query to the open list of the opposite frontier to check for the existence of a node with the same state as the newly generated node $n$. This is trivially implemented in \UniHS, as it only entails checking if the state of $n$ is the goal. Similarly, in standard \BiHS, 
when the open lists are stored in memory, ISD can be efficiently performed using a hash table or a direct-access table where each item can be accessed in constant time. 

However, when buckets are stored on disk, ISD can lead to frequent I/O calls every time a node is generated or when the cache of generated nodes is filled. Therefore, \pembfs employs DSD as suggested by \citet{external2016sturtevant}. In DSD, solutions are identified during the expansion phase once all states of the bucket that was chosen to be expanded are already loaded into memory and stored in a hash table. Following that, closed buckets from the opposite direction are loaded in segments and compared against the hash table of expanded nodes in an effort to identify a solution. It's important to highlight that only relevant buckets need to be loaded; for instance, if the identifiers of the bucket of expanded nodes include \hF\ and \hB\ values, only closed buckets corresponding to the same $h$-values need to be considered. We note again that when the expansion and solution detection are done, all threads are synchronized.

It is important to acknowledge that the benefits of DSD come with certain trade-offs. First, since solutions are detected at a later stage, some nodes might be expanded which could have been avoided with ISD. Furthermore, search bounds leveraging information across the minimal edge cost (often denoted as $\epsilon$) cannot be employed with DSD. This limitation arises because these bounds rely on ISD to improve the lower bound ($LB$) on the solution cost (we refer the reader to \citet{external2016sturtevant} for more details).

\section{Parallel External-Memory BAE*}

We next describe the implementation details needed for obtaining a PEM variant of the \bae algorithm (PEM-BAE*).

\noindent\textbf{Direction, Prioritization, and Lower-bound.} The prioritization of buckets relies on the \bae priority function (Eq.\ref{eq:BAE}).  The lower bound is determined by the $b$-bound (Eq.\ref{eq:BAE-U}). Additionally, the direction selection policy alternates between directions; a seemingly simplistic approach that has proven to be as effective as more intricate policies~\citep {alcazar20unifying}. In PEM, this means that the two search sides take turns in loading buckets into memory and expanding them.

\noindent\textbf{Bucket Structure.} In PEM-BAE*, we classify nodes into buckets based on their $g_D$-, $h_F$-, and $h_B$-values. 
This bucket structure offers several advantages. First, it encapsulates all the essential information needed for computing the \bae priority function. Consequently, during each expansion cycle, only a single bucket needs to be loaded into memory, enhancing efficiency. Second, the $g$-value, serving as one of the bucket identifiers, can sometimes further reduce the number of closed buckets scanned during duplicate detection (as discussed in Section~\ref{ssec:duplicate_detection}). Finally, as the classification is based on three values, the resulting buckets tend to remain relatively small. Notably, when buckets become too large for memory, a hash value can be introduced as an additional identifier but such a scenario did not arise in our experiments.
A possible limitation of this approach is the potential for significant imbalances in the distribution of nodes across buckets. This bucket size imbalance could impede runtime performance, as we explore in our experiments below.

\noindent\textbf{Solution Detection.} PEM-BAE* uses DSD.

\section{Experimental Results} \label{sec:experiments}

We performed experiments with \PEMBFS on the 15- and 24-sliding-tile puzzles (STP) and 4-peg Towers of Hanoi (ToH4). All experiments were executed on 2 Intel Xeon Gold 6248R Processor 24-Core 3.0GHz, 192 GB of 3200MHz DDR4 RAM, and 100TB SSD for the external memory. By default, all parallel algorithms were assessed utilizing all 96 available virtual threads (with 48 physical cores). Nevertheless, we have also conducted an ablation study to investigate algorithm performance while varying the number of threads.

We tested the following algorithms besides PEM-BAE*. First, we instantiated both \astar and \MM within the \pembfs framework, resulting in \PEMASTAR and \pemm, respectively. Similarly to \astar, \PEMASTAR consistently expands nodes in the forward direction, employs the minimal $f$-value in \open as a lower bound for the solution cost ($LB$), and expands a g-h bucket with the minimal $f$-value during each expansion cycle. \PEMASTAR~uses the low-$g$-first  tie-breaking as detailed in Section~\ref{ssec:dsbplb}. \PEMASTAR incorporates ISD, checking for solution upon node generation, where the \PEMBAE~and \pemm~employ DSD, checking for solutions before node expansions.
As \BiHS search algorithms explore from both the \goal and the \start states, it is crucial to ensure that any potential advantage is not merely a consequence of asymmetries, which causes the search tree from one side to be much smaller. Such asymmetries could also be leveraged in a unidirectional search from \goal to \start. To address this concern, we executed a reverse variant of \PEMASTAR, denoted as \RPEMSTAR, where the search is conducted from \goal to \start.

\PEMM uses the direction-selection, lower-bound, and prioritizations of \mm while using the lower-$g$-first tie-breaking. \pemm uses the $g$-value and priority value (Eq.~\ref{eq:MM}) as bucket identifiers. As a \BiHS, \pemm adopts DSD as part of its operational strategy.

For comparison, we have also implemented Asynchronous Parallel \IDAStar (\AIDAStar), \citep{reinefeld1994work}). \AIDAStar is a parallelized adaptation of IDA* which conducts a breadth-first search to a predetermined depth. The resulting frontier is subsequently distributed among all threads. Additionally, we've evaluated the reverse version of \AIDAStar, referred to as \AIDARStar. Lastly, we evaluated the standard versions of \astar and \bae, without parallelization or external memory.

\begin{figure}[tb]
  \centering
  \begin{minipage}[b]{0.45\linewidth}
    \centering
    \includegraphics[width=\linewidth]{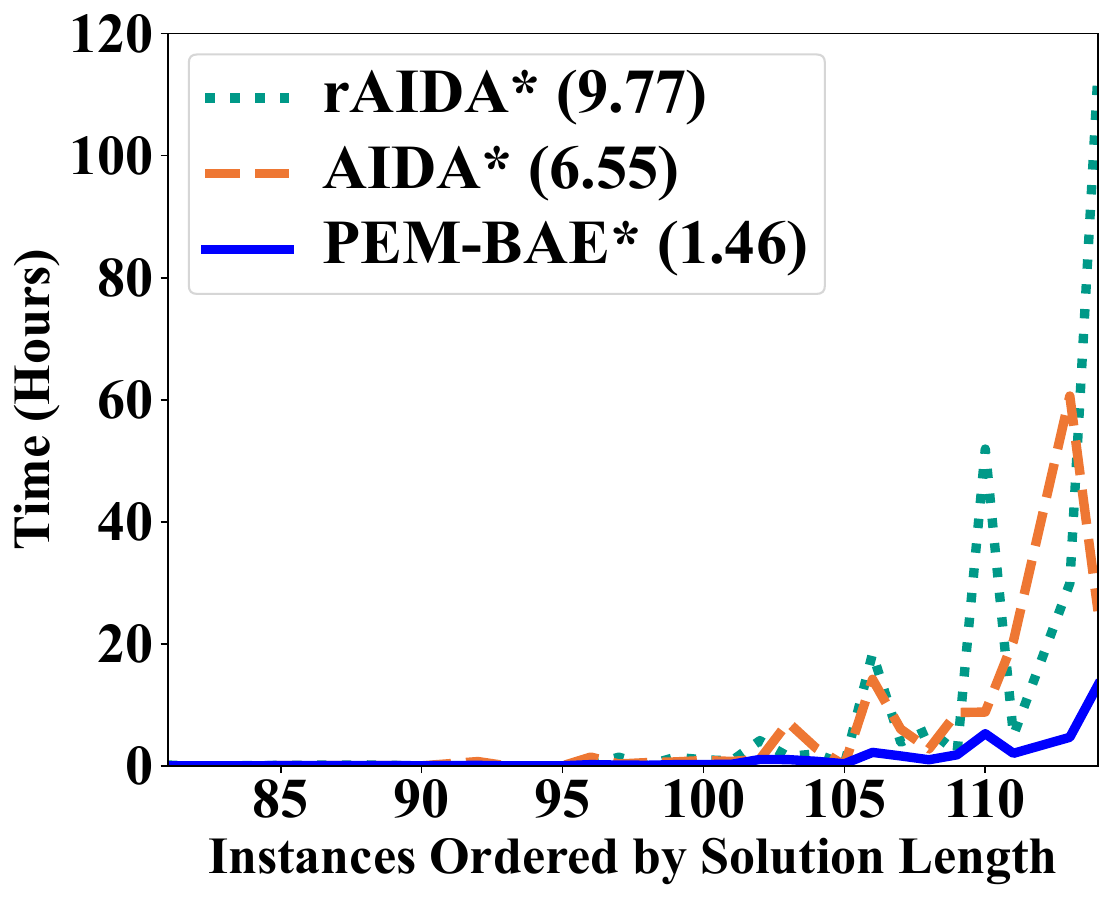}
  \end{minipage}
  \hfill
  \begin{minipage}[b]{0.45\linewidth}
    \centering
    \includegraphics[width=\linewidth]{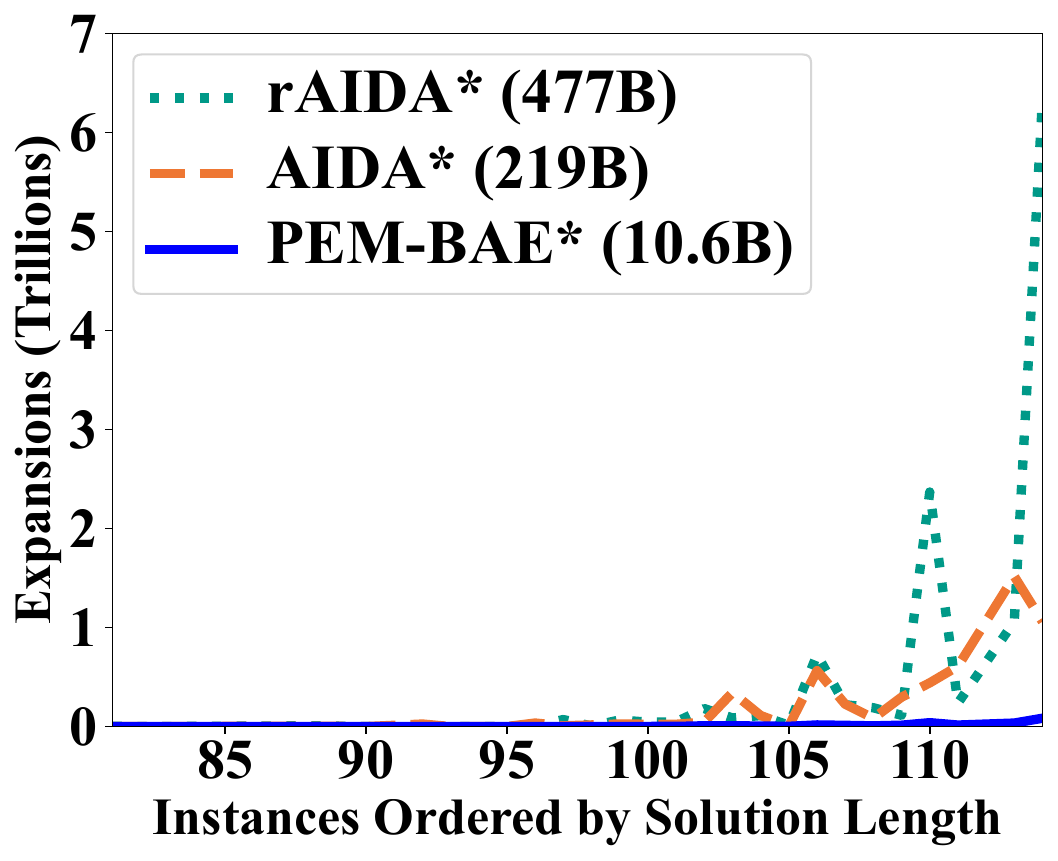}
  \end{minipage}
  \caption{24-puzzle results: (left) runtime, (right) node expansions}
    \label{fig:24stp}
\end{figure}

\subsection{15-Puzzle}
\label{ssec:stp4}

\begin{table}[tb]
\centering
\setlength{\tabcolsep}{3pt}
\begin{tabular}{l|rr|rr|}
\cline{2-5}
 & \multicolumn{2}{c|}{\textbf{MD}} & \multicolumn{2}{c|}{\textbf{PDB}} \\
 & \multicolumn{1}{c|}{Time} & \multicolumn{1}{c|}{Expansions} & \multicolumn{1}{c|}{Time} & \multicolumn{1}{c|}{Expansions} \\ 
 \hline 
\multicolumn{5}{|c|}{All instances}\\
\hline
\multicolumn{1}{|l|}{\PIDA} & \multicolumn{1}{r|}{3.45} & 451,421,959 & \multicolumn{1}{r|}{0.43} & 7,762,927 \\
\multicolumn{1}{|l|}{\RPIDA} & \multicolumn{1}{r|}{{\bf 2.44}} & 335,167,556 & \multicolumn{1}{r|}{{\bf 0.37}} & 6,118,084 \\ \hline
\multicolumn{1}{|l|}{\PEMASTAR} & \multicolumn{1}{r|}{102.33} & 56,542,721 & \multicolumn{1}{r|}{2.01} & 2,724,974 \\
\multicolumn{1}{|l|}{\RPEMSTAR} & \multicolumn{1}{r|}{84.38} & 43,451,519 & \multicolumn{1}{r|}{1.85} & 2,302,668 \\ \hline
\multicolumn{1}{|l|}{PEM-MM} & \multicolumn{1}{r|}{16.49} & 26,771,047 & \multicolumn{1}{r|}{5.2} & 2,572,780 \\
\multicolumn{1}{|l|}{\PEMBAE} & \multicolumn{1}{r|}{6.11} & {\bf 3,113,271} & \multicolumn{1}{r|}{3.06} & {\bf 626,440} \\ \hline \hline
\multicolumn{1}{|l|}{A*} & \multicolumn{1}{r|}{56.88} & 15,549,689 & \multicolumn{1}{r|}{2.22} & 615,155 \\
\multicolumn{1}{|l|}{BAE*} & \multicolumn{1}{r|}{\textbf{10.13}} & \textbf{2,707,414} & \multicolumn{1}{r|}{\textbf{1.76}} & \textbf{453,988} \\
\multicolumn{1}{|l|}{IDA*} & \multicolumn{1}{r|}{51.82} & 242,460,834 & \multicolumn{1}{r|}{2.67} & 3,456,177 \\
\hline  \hline
\multicolumn{5}{|c|}{The 10 hard instances: 3, 15, 17, 32, 49, 56, 60, 66, 82, 88}\\
\hline 
\multicolumn{1}{|l|}{\PIDA} & \multicolumn{1}{r|}{22.18} & 2,943,505,999 & \multicolumn{1}{r|}{2.13} & 46,314,389 \\
\multicolumn{1}{|l|}{\RPIDA} & \multicolumn{1}{r|}{16.67} & 2,695,821,070 & \multicolumn{1}{r|}{\textbf{1.93}} & 41,047,358 \\ \hline 
\multicolumn{1}{|l|}{\PEMASTAR} & \multicolumn{1}{r|}{901.19} & 350,840,875 & \multicolumn{1}{r|}{7.8} & 17,124,704 \\
\multicolumn{1}{|l|}{\RPEMSTAR} & \multicolumn{1}{r|}{786.58} & 308,829,220 & \multicolumn{1}{r|}{6.67} & 14,371,919 \\ \hline
\multicolumn{1}{|l|}{PEM-MM} & \multicolumn{1}{r|}{74.14} & 165,459,580 & \multicolumn{1}{r|}{13.07} & 14,989,610 \\
\multicolumn{1}{|l|}{\PEMBAE} & \multicolumn{1}{r|}{\textbf{13.31}} & \textbf{15,749,202} & \multicolumn{1}{r|}{6.05} & \textbf{3,199,891} \\ \hline \hline
\multicolumn{1}{|l|}{A*} & \multicolumn{1}{r|}{378.41} & 98,596,826 & \multicolumn{1}{r|}{13.63} & 3,636,711 \\
\multicolumn{1}{|l|}{BAE*} & \multicolumn{1}{r|}{\textbf{56.69}} & \textbf{13,865,491} & \multicolumn{1}{r|}{\textbf{10.33}} & \textbf{2,451,979}\\
\multicolumn{1}{|l|}{IDA*} & \multicolumn{1}{r|}{368.81} & 1,731,811,022 & \multicolumn{1}{r|}{16.79} & 22,069,583 \\\hline
\end{tabular}
\caption{15-puzzle Results. Avg. time in seconds.}
\label{tab:stp4_all}

\end{table}

We first experimented on Korf's 100 random instances of the 15-puzzle~\cite{korf1985depth}. This domain is relatively compact ($13^{10}$ states) and could be solved without external memory. However, its size enables a comprehensive comparison of all algorithms across different heuristics before moving to larger domains. For heuristics, we used  Manhattan Distance (labeled MD) and a 3-4-4-4 additive pattern database~\citep{felner2004additive}) (labeled PDB).
To construct the PDB, we divided the puzzle into four squares, one for each corner, with each pattern also including the blank.
The average runtime (in seconds) and the number of node expansions are presented in \Cref{tab:stp4_all} (top).

Naturally, using the PDB heuristic substantially reduced both the time and node expansions for all algorithms when compared to using the MD heuristic. The \AIDAStar variants expanded the largest number of nodes but exhibited the fastest overall runtime. Due to their DFS nature, the \AIDAStar variants do not store nodes in memory, let alone external memory, nor do they involve sorting nodes, as typically done by best-first search algorithms.  Thus, they have the smallest time overhead per node in comparison with all other algorithms.
Among the PEM algorithms, both \BiHS algorithms outperformed the \UniHS algorithms in terms of node expansions and runtime, when using the MD heuristic. Notably, \PEMBAE significantly outperformed {\em all} PEM or iterative deepening algorithms in terms of node expansions by an order of magnitude. This finding is consistent with prior studies that compared \bae to \astar~\cite{alcazar20unifying, siag2023front}, underscoring the advantages of harnessing the consistency of heuristics, as demonstrated by the performance of both \astar and \bae as presented in the table.
Even when employing the PDB heuristic, \PEMBAE maintained a significant edge over all other PEM or iterative deepening algorithms in terms of node expansions. However, it exhibited slower runtime than \PEMASTAR and \RPEMSTAR, a phenomenon we address in Section~\ref{ssec:NPS}. Furthermore, despite its parallelism, \PEMBAE was slower than both \astar and \bae when using the PDB heuristic. As anticipated, in scenarios where the problems are relatively simple to solve, the overhead incurred by utilizing external memory and parallelization serves to decelerate the search process rather than expedite it.
\begin{figure}[tb]
  \centering
  \begin{minipage}[b]{0.45\linewidth}
    \centering
    \includegraphics[width=\linewidth]{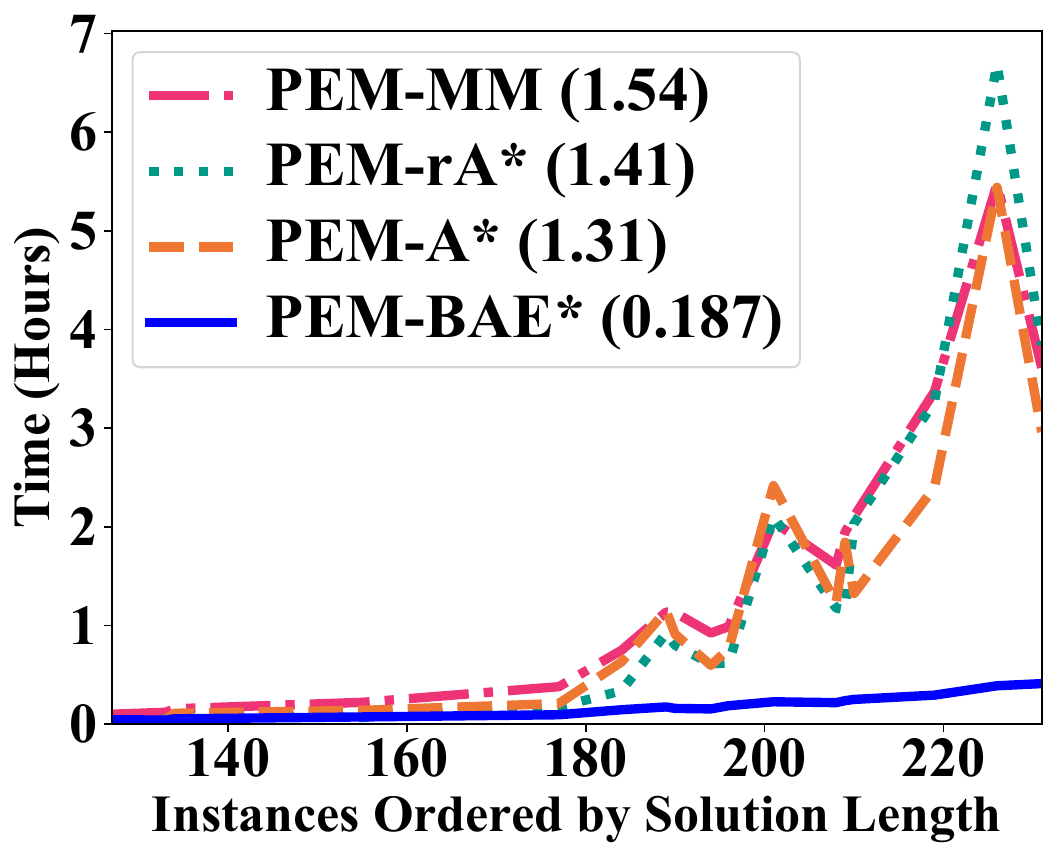}
  \end{minipage}
  \hfill
  \begin{minipage}[b]{0.45\linewidth}
    \centering
    \includegraphics[width=\linewidth]{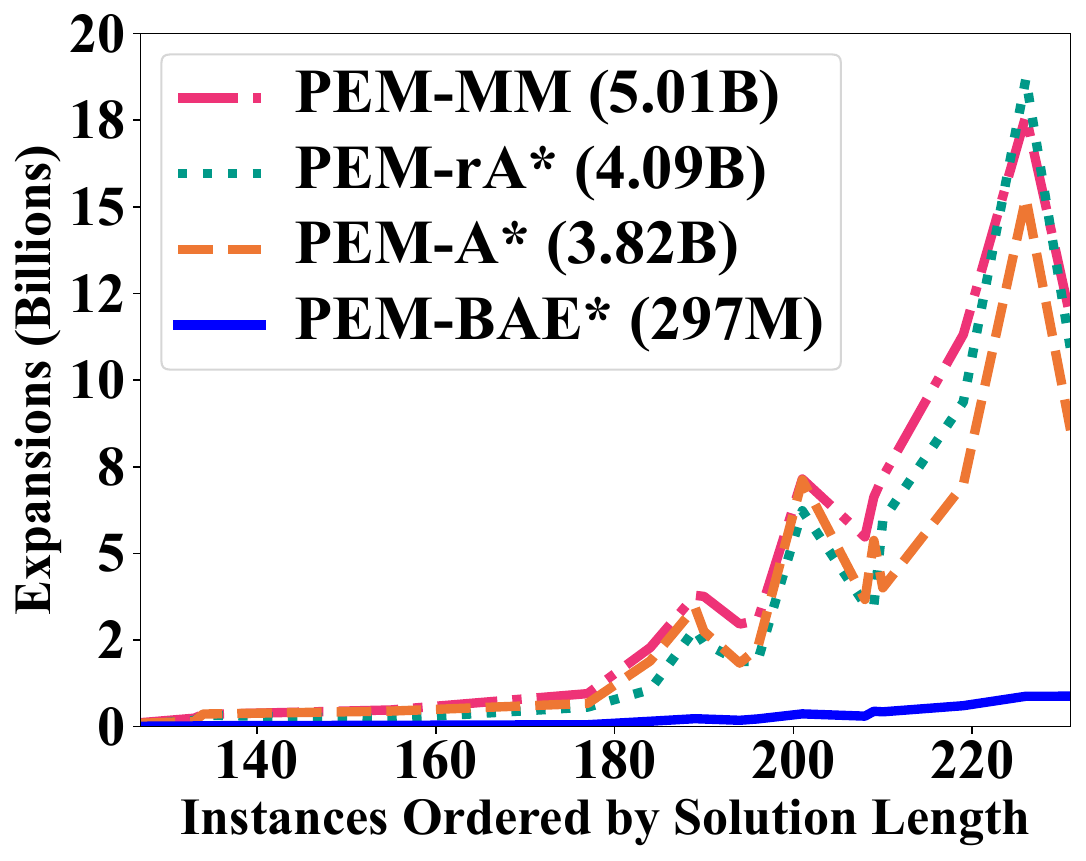}
  \end{minipage}
  \caption{ToH4 16+4 results: (left) runtime, (right) node expansions.}
    \label{fig:toh}
\end{figure}

\begin{figure*}
  \centering
  \begin{minipage}[b]{0.28\textwidth}
    \centering
    \includegraphics[width=\textwidth,valign=t]{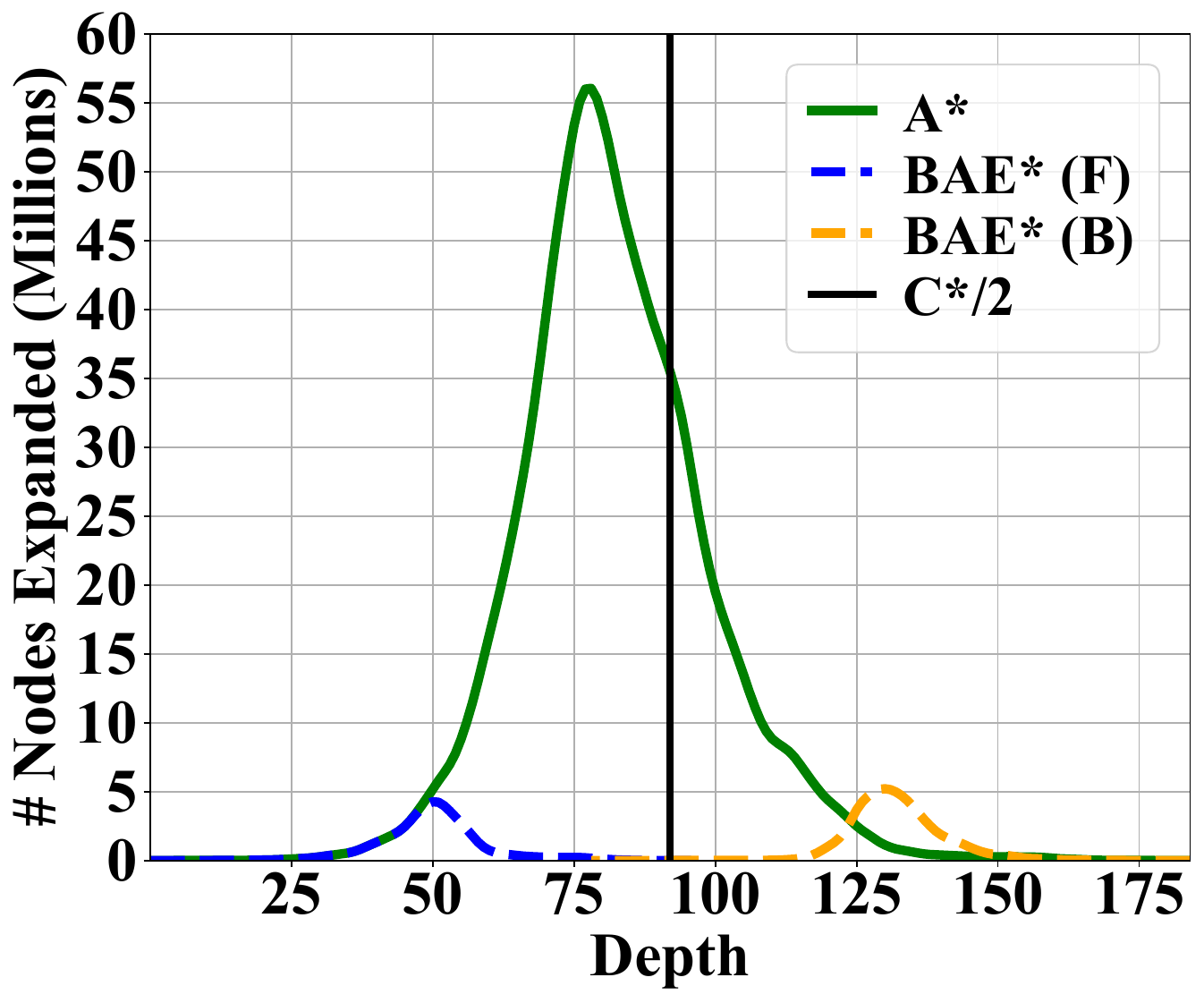}
    \caption{Distribution of node expansions in ToH4 instance with solution length of 184}
    \label{fig:distribution}
  \end{minipage}%
  \hfill
  \begin{minipage}[b]{0.32\textwidth}
    \centering
    \includegraphics[width=\textwidth,valign=c]{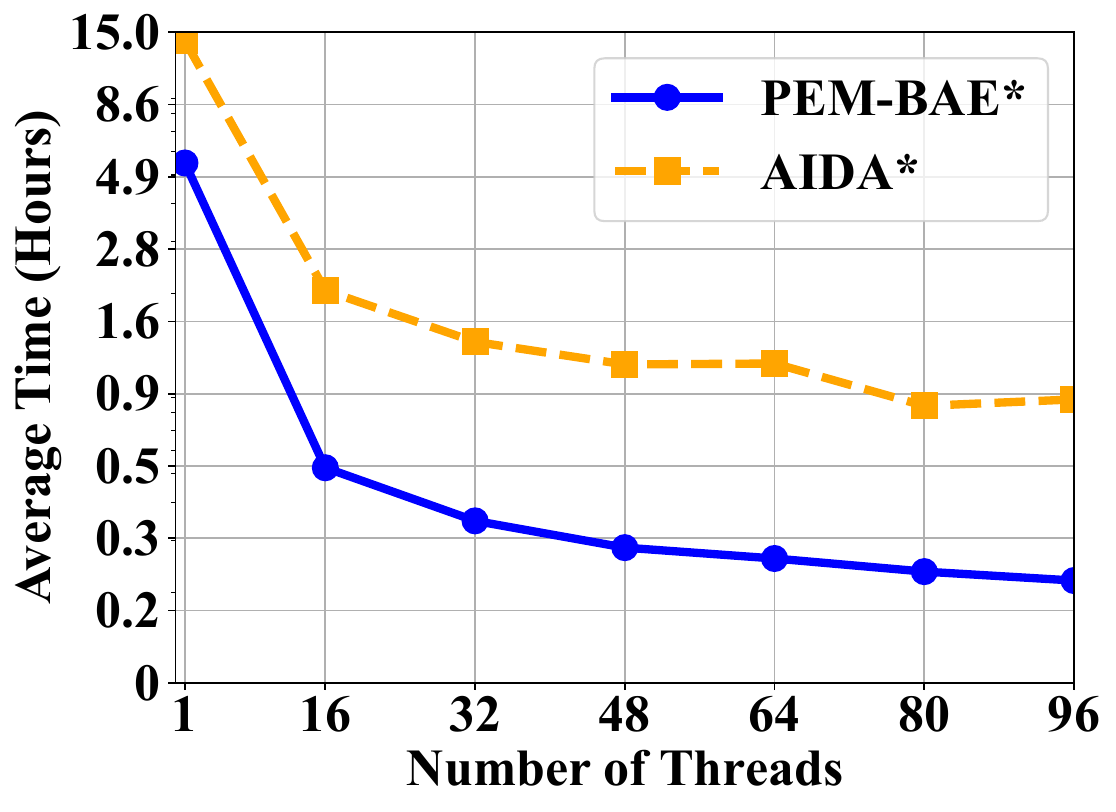}
    \caption{Avg. runtime as a function of number of threads in 24-Puzzle}
    \label{fig:ablation}
  \end{minipage}%
  \hfill
  \begin{minipage}[b]{0.4\textwidth}
    \centering
\resizebox{\linewidth}{!}{
\begin{adjustbox}{valign=c}
\begin{tabular}{l|r|r|r|r|}
\cline{2-5}
 & \PEMBAE & \PEMM & \PEMASTAR & \RPEMSTAR \\ \hline
\multicolumn{1}{|l|}{15-STP$_{\text{MD}}$} & 509,537 & \textbf{1,623,472} & 552,553 & 514,950 \\
\multicolumn{1}{|l|}{15-STP$_{\text{PDB}}$} & 204,719 & 494,765 & \textbf{1,355,708} & 1,244,685 \\ \hline
\multicolumn{1}{|l|}{15-STP$^H_{\text{MD}}$} & 1,183,261 & \textbf{2,231,718} & 389,308 & 392,623 \\
\multicolumn{1}{|l|}{15-STP$^H_{\text{PDB}}$} & 528,908 & 1,146,871 & \textbf{2,195,475} & 2,154,710 \\ \hline
\multicolumn{1}{|l|}{TOH4} & 439,122 & \textbf{908,970} & 811,984 & 808,777 \\ \hline \hline
\multicolumn{1}{|l|}{Average} & 573,109 & \textbf{1,281,159} & 1,061,006 & 1,023,149\\ \hline
\end{tabular}
\end{adjustbox}
}

\caption{Avg. nodes expanded per second; $^H$ indicates the hard instance in STP.}
\label{tb:nps}

  \end{minipage}
\end{figure*}

\Cref{tab:stp4_all} (bottom) provides results on the ten most difficult instances (with largest solution cost). 
In these instances, the superiority of \PEMBAE is more pronounced in terms of node expansions, and it also outperforms \PEMASTAR and \RPEMSTAR in terms of runtime. Additionally, \PEMBAE outperformed both \PIDA and \RPIDA with MD. Although \PEMBAE was still slower than \PIDA and \RPIDA with PDB heuristic, the performance gap is comparatively narrower.

As the problems become more challenging or, conversely, when the heuristic weakens, the performance of \PEMBAE improves relative to other algorithms. This is demonstrated next on the 24-Puzzle, marking the first evaluation of \BiHS algorithms for this large domain.

\subsection{24-Puzzle}

We experimented with the 50 24-puzzle problems of \citet{korf2002disjoint}, using a 6+6+6+6 additive PDB heuristic coupled with its reflection about the main diagonal~\citep{felner2004additive}. Given the immense size of these 24-puzzle problems and the extensive computation time they demand, assessing all algorithms becomes impractical. Moreover, the memory demands for tackling these problems are substantial, making it impractical to execute standard (in-memory) \astar and \bae algorithms.
Thus, we compared \pembfs with the AIDA* variants. \Cref{fig:24stp} illustrates the runtime (left) and the number of expanded nodes (right) for each instance. 
Instances are sorted solution length in ascending order, which roughly indicates the problem's difficulty. For instances with the same solution length, we averaged the results. The plot legends also display the average runtime and expansions for each algorithm across all instances, shown in parentheses.

In general, \PEMBAE performs the best in both node expansions and runtime. On average, \PEMBAE expands only $4.4\%$ of the nodes expanded by \PIDA and runs $4.5$ times faster. These findings align with the observed trend in the 15-puzzle, indicating that on challenging problems, \PEMBAE outperforms \UniHS algorithms even when equipped with state-of-the-art (or near state-of-the-art) heuristics. The average disk space consumption of \PEMBAE for the 5x5 STP instances was 653GB, with a peak of 4TB. This underscores the impracticality of employing in-memory algorithms (such as \astar and \bae) for tackling these challenging problems.

\subsection{4-peg Towers of Hanoi}

In TOH4, we examined 20 random start and goal pairs with 20 disks, utilizing a 16+4 additive PDB heuristic~\citep{felner2004additive}. In this domain, numerous cycles exist, posing a challenge for algorithms that lack duplicate detection, as already noted by \citet{felner2004additive}. This issue is so severe that neither \PIDA nor \RPIDA could solve a single problem even after running for days. Consequently, we only compared \PEMBAE, \PEMASTAR, \RPEMSTAR, and PEM-MM.

The results, presented in \Cref{fig:toh}, highlight a significant performance gap between \PEMBAE and the other algorithms.
On average, \PEMBAE runs 7 times faster than its \UniHS counterparts and expands a factor of 12.9  fewer nodes. Notably, \PEMM was approximately 1.17 times slower than both \PEMASTAR and \RPEMSTAR, and it expanded more nodes than both of them.

\subsection{Analyzing Previous Conjectures}

In a previous analysis of bidirectional search~\cite{barker2015limitations}, it was suggested that in a unidirectional search ``if the majority [{\em of nodes}] are expanded at shallower depth than the solution midpoint [$\CStar/2$] then [...] a bidirectional heuristic search would expand more nodes than a unidirectional heuristic search.'' . The analysis for this claim pre-dates our current understanding of \BiHS. But, since we have the data it is worthwhile to evaluate the validity of this claim.

In this context, in our experiments we found that in 19 of 20 Towers of Hanoi instances and all 15-puzzle instances when using the PDB, the unidirectional algorithm (\PEMASTAR) expanded the majority of states prior to the solution midpoint. Yet, in all of these problems \PEMBAE expanded fewer node than \PEMASTAR. Thus, the previous analysis does not hold for \PEMBAE on these problems. This analysis for a representative ToH4 instance is presented in Figure~\ref{fig:distribution}. It is a matter for future work to analyze these claims in more depth and to consider whether or how to revise them to be more accurate.

\subsection{Analyzing Expansion-per-second Ratios} \label{ssec:NPS}

The Table in \Cref{tb:nps} presents the number of nodes expanded per second (NPS) by different PEM algorithms on the domains in which all PEM algorithms were evaluated, namely 15-STP and TOH4. While these algorithms differ in how they choose their search direction, decide on which bucket to expand, and terminate the search, these differences are not expected to significantly impact NPS. Two factors, however, have the potential to affect NPS.
First, \BiHS algorithms perform DSD, while \UniHS algorithms perform ISD. DSD generally requires more computational effort, though a profiling analysis revealed that this additional time was negligible.
Second, differences in bucket structure can affect the number of buckets and their sizes, affecting the I/O time of the algorithms. The results show significant variations in NPS among the algorithms. Despite its relative strength in nodes and in time, \PEMBAE exhibited the worst (smallest) NPS overall, suggesting that an alternative bucket structure might further improve its advantages over the other algorithms.

\subsection{Ablation Study on the Number of Threads} \label{ssec:Ablation}

To assess the thread utilization of \PEMBAE, we compare its performance against \AIDAStar on the 24-STP while varying number of threads. Specifically, we conducted experiments using $1$, $16$, $32$, $48$, $64$, $80$, and $96$ (virtual) threads on a subset of problems (problems 4, 36, 45, 48) with an intermediate solution cost, $C^*=100$ .

\Cref{fig:ablation} shows average runtimes for various thread configurations, with a logarithmic $y$-axis. As observed previously, \PEMBAE surpasses \AIDAStar performance with 96 threads and consistently outperforms it across varying thread counts. 
As anticipated, adding more threads yields diminishing returns. \PEMBAE reduced its runtime by a factor of 11 when transitioning from 1 thread to 16 threads, whereas \AIDAStar improved by a factor of 7. Beyond 16 threads, runtime reduction becomes smaller, decreasing only by a factor of 2 for both algorithms when transitioning from 16 to 96 threads. This reduced gain can be attributed to memory access required for obtaining heuristic values, imbalanced subtrees and last-layer expansions for \AIDAStar, constrained I/O parallelization (relative to the thread count), imbalanced bucket sizes, and locking overhead for \pembfs.

\section{Conclusions and Future Work}
We presented \pembfs, a parallel external-memory (PEM) \BiHS framework for single-target search in undirected, uniformly weighted search graphs, which was used to create a PEM variant of \bae (\PEMBAE). Our empirical evaluations show that \PEMBAE outperforms \UniHS algorithms both in runtime and node expansions, even with well-informed heuristics. These findings challenge the conjecture put forth by \citet{barker2015limitations}, suggesting that \BiHS algorithms would not significantly surpass \UniHS or bidirectional brute-force search. Further theoretical study is necessary to analyze our results in relation to this conjecture, as well as to other theoretical comparisons between \BiHS and \UniHS algorithms~\citep{holte2017mm, sturtevant2020predicting}.

Future research could also address NPS differences among algorithms by exploring dynamic bucket sizes and other approaches that relax the best-first assumption, aiming to achieve more balanced buckets~\citep{hatem2018solving}.

%%%%%%%%%%%%%%%%%%%%%%%%%%%%%%%%%%%%%%%%%%%%%%%%%%%%%%%%%%%%%%%%%%%%%%%%

%%% Use this environment to include acknowledgements (optional).
%%% This will be omitted in doubleblind mode.

\begin{ack}
This work was supported by the Israel Science Foundation (ISF)
grant \#909/23 awarded to Shahaf Shperberg and Ariel Felner,  by Israel's Ministry of Innovation, Science and Technology (MOST) grant \#1001706842, awarded to Shahaf Shperberg, and by United States-Israel Binational Science Foundation (BSF) grant \#2021643 awarded to Ariel Felner. This work was also partially funded by the Canada CIFAR AI Chairs Program. We acknowledge the support of the National Sciences and Engineering Research Council of Canada (NSERC).
\end{ack}

%%%%%%%%%%%%%%%%%%%%%%%%%%%%%%%%%%%%%%%%%%%%%%%%%%%%%%%%%%%%%%%%%%%%%%%%

%%% Use this command to include your bibliography file.

\bibliography{main}

\newpage
\appendix
\section{Code}
The code of the paper was uploaded to GitHub and can be found here: https://github.com/SPL-BGU/PEM-BiHS

\section{Results Tables}
Tables \ref{app:tab:stp5_first}, \ref{app:tab:stp_second}, and \ref{app:tab:toh} present the full results summarized in the paper for 24-Sliding Tile Puzzle and 4-Peg Towers of Hanoi respectivley. The 24-STP instances match the instances described in Table 2 of Korf~\cite{korf2002disjoint}. The ToH instances correspond to random ones generated (a full list appears in the code repo).
The columns include the following: instance number, the algorithm executed, the number of nodes expanded, the number of nodes generated, time in seconds, and the solution cost.

\begin{table}[htb]
\begin{adjustbox}{width=0.93\columnwidth}
\begin{tabular}{|l|l|r|r|r|r|}
\hline
\multicolumn{1}{|c|}{\textbf{Instance}} & \multicolumn{1}{c|}{\textbf{Algorithm}} & \multicolumn{1}{c|}{\textbf{Expanded}} & \multicolumn{1}{c|}{\textbf{Generated}} & \multicolumn{1}{c|}{\textbf{Elapsed}} & \multicolumn{1}{c|}{\textbf{Solution}} \\
\hline
0 & PEM-BAE* & 69,230,832 & 220,879,373 & 48 & 95 \\
0 & AIDA* & 703,361,151 & 2,230,288,795 & 60 & 95 \\
0 & rAIDA* & 298,891,181 & 962,050,469 & 21 & 95 \\
\hline
1 & PEM-BAE* & 4,268,907,890 & 13,630,651,253 & 1,994 & 96 \\
1 & AIDA* & 106,013,766,980 & 341,876,855,982 & 15,411 & 96 \\
1 & rAIDA* & 11,115,669,857 & 35,192,151,412 & 1,383 & 96 \\
\hline
2 & PEM-BAE* & 1,881,452,125 & 6,079,830,316 & 870 & 97 \\
2 & AIDA* & 15,497,174,429 & 50,045,721,079 & 2,207 & 97 \\
2 & rAIDA* & 118,194,520,400 & 386,919,173,185 & 8,559 & 97 \\
\hline
3 & PEM-BAE* & 486,925,192 & 1,564,458,403 & 241 & 98 \\
3 & AIDA* & 10,499,950,390 & 34,166,927,429 & 916 & 98 \\
3 & rAIDA* & 8,355,167,278 & 26,527,013,652 & 647 & 98 \\
\hline
4 & PEM-BAE* & 562,259,551 & 1,784,446,623 & 266 & 100 \\
4 & AIDA* & 6,251,407,352 & 19,828,603,739 & 429 & 100 \\
4 & rAIDA* & 4,829,480,771 & 15,612,945,465 & 404 & 100 \\
\hline
5 & PEM-BAE* & 3,060,348,563 & 9,778,084,836 & 1,408 & 101 \\
5 & AIDA* & 39,864,885,473 & 127,574,030,552 & 4,617 & 101 \\
5 & rAIDA* & 126,120,370,265 & 402,272,063,608 & 7,545 & 101 \\
\hline
6 & PEM-BAE* & 2,645,665,630 & 8,485,295,970 & 1,206 & 104 \\
6 & AIDA* & 85,638,897,490 & 272,142,756,965 & 6,424 & 104 \\
6 & rAIDA* & 69,732,574,150 & 224,503,116,339 & 5,088 & 104 \\
\hline
7 & PEM-BAE* & 6,341,150,974 & 20,222,934,040 & 2,922 & 108 \\
7 & AIDA* & 25,961,585,022 & 82,385,607,008 & 6,254 & 108 \\
7 & rAIDA* & 133,373,261,394 & 427,651,175,162 & 13,143 & 108 \\
\hline
8 & PEM-BAE* & 27,212,933,333 & 86,599,962,301 & 13,504 & 113 \\
8 & AIDA* & 776,302,510,731 & 2,460,969,417,979 & 59,157 & 113 \\
8 & rAIDA* & 1,779,553,689,438 & 5,629,539,624,191 & 189,967 & 113 \\
\hline
9 & PEM-BAE* & 84,643,679,974 & 268,176,125,953 & 47,688 & 114 \\
9 & AIDA* & 1,063,877,057,715 & 3,330,671,280,447 & 91,349 & 114 \\
9 & rAIDA* & 6,341,997,459,809 & 19,959,427,954,921 & 409,236 & 114 \\
\hline
10 & PEM-BAE* & 22,043,034,863 & 70,424,617,713 & 10,265 & 106 \\
10 & AIDA* & 1,790,268,187,428 & 5,712,762,291,643 & 163,722 & 106 \\
10 & rAIDA* & 338,961,713,914 & 1,064,576,618,761 & 25,734 & 106 \\
\hline
11 & PEM-BAE* & 7,166,972,315 & 22,552,629,538 & 3,292 & 109 \\
11 & AIDA* & 163,240,926,083 & 513,056,203,546 & 15,620 & 109 \\
11 & rAIDA* & 17,864,928,517 & 55,236,406,222 & 1,047 & 109 \\
\hline
12 & PEM-BAE* & 810,360,812 & 2,560,814,092 & 372 & 101 \\
12 & AIDA* & 3,326,240,519 & 10,389,525,447 & 261 & 101 \\
12 & rAIDA* & 5,363,614,647 & 17,014,790,519 & 562 & 101 \\
\hline
13 & PEM-BAE* & 15,722,616,097 & 50,530,082,815 & 7,416 & 111 \\
13 & AIDA* & 610,358,362,103 & 1,974,350,435,375 & 73,723 & 111 \\
13 & rAIDA* & 237,342,857,210 & 748,028,335,178 & 17,874 & 111 \\
\hline
14 & PEM-BAE* & 6,546,441,457 & 21,088,442,538 & 3,245 & 103 \\
14 & AIDA* & 121,732,630,600 & 394,521,427,843 & 13,083 & 103 \\
14 & rAIDA* & 96,099,514,702 & 309,679,093,803 & 6,681 & 103 \\
\hline
15 & PEM-BAE* & 512,297,135 & 1,649,387,410 & 250 & 96 \\
15 & AIDA* & 2,210,852,579 & 7,109,629,829 & 198 & 96 \\
15 & rAIDA* & 12,104,098,103 & 39,207,380,078 & 1,118 & 96 \\
\hline
16 & PEM-BAE* & 21,572,731,051 & 67,876,835,431 & 9,793 & 109 \\
16 & AIDA* & 431,775,179,640 & 1,355,237,512,487 & 47,360 & 109 \\
16 & rAIDA* & 210,983,589,712 & 655,921,859,058 & 14,793 & 109 \\
\hline
17 & PEM-BAE* & 39,843,274,192 & 126,680,194,817 & 18,998 & 110 \\
17 & AIDA* & 442,310,527,540 & 1,398,526,796,413 & 31,757 & 110 \\
17 & rAIDA* & 2,374,565,687,766 & 7,512,383,106,222 & 186,946 & 110 \\
\hline
18 & PEM-BAE* & 10,121,572,991 & 31,893,504,326 & 4,404 & 106 \\
18 & AIDA* & 132,965,680,106 & 413,222,129,725 & 9,021 & 106 \\
18 & rAIDA* & 76,199,307,271 & 241,731,795,548 & 4,883 & 106 \\
\hline
19 & PEM-BAE* & 1,334,088,442 & 4,267,729,893 & 584 & 92 \\
19 & AIDA* & 67,664,490,886 & 215,174,841,090 & 6,093 & 92 \\
19 & rAIDA* & 1,110,753,209 & 3,615,804,097 & 83 & 92 \\
\hline
\end{tabular}
\end{adjustbox}
\caption{24-STP Instances 0-20}
\label{app:tab:stp5_first}
\end{table}

\begin{table}[htb]
\begin{adjustbox}{width=0.93\columnwidth}
\begin{tabular}{|l|l|r|r|r|r|}
\hline
\multicolumn{1}{|c|}{\textbf{Instance}} & \multicolumn{1}{c|}{\textbf{Algorithm}} & \multicolumn{1}{c|}{\textbf{Expanded}} & \multicolumn{1}{c|}{\textbf{Generated}} & \multicolumn{1}{c|}{\textbf{Elapsed}} & \multicolumn{1}{c|}{\textbf{Solution}} \\
\hline
20 & PEM-BAE* & 9,527,278,036 & 30,356,569,236 & 4,188 & 103 \\
20 & AIDA* & 565,380,369,670 & 1,795,962,930,341 & 38,295 & 103 \\
20 & rAIDA* & 78,458,996,064 & 249,254,041,120 & 4,581 & 103 \\
\hline
21 & PEM-BAE* & 214,615,450 & 686,779,942 & 110 & 95 \\
21 & AIDA* & 963,175,022 & 3,112,348,975 & 78 & 95 \\
21 & rAIDA* & 1,135,413,445 & 3,640,912,326 & 70 & 95 \\
\hline
22 & PEM-BAE* & 4,029,225,810 & 12,910,448,751 & 1,734 & 104 \\
22 & AIDA* & 80,547,390,480 & 257,197,389,109 & 11,128 & 104 \\
22 & rAIDA* & 18,303,074,822 & 58,970,515,060 & 1,181 & 104 \\
\hline
23 & PEM-BAE* & 8,301,166,057 & 26,225,686,035 & 3,704 & 107 \\
23 & AIDA* & 169,310,755,886 & 530,726,996,648 & 21,803 & 107 \\
23 & rAIDA* & 198,425,749,734 & 632,052,577,237 & 13,928 & 107 \\
\hline
24 & PEM-BAE* & 65,808,216 & 214,259,822 & 47 & 81 \\
24 & AIDA* & 234,082,829 & 757,591,010 & 33 & 81 \\
24 & rAIDA* & 3,415,426,809 & 11,218,683,998 & 478 & 81 \\
\hline
25 & PEM-BAE* & 3,002,690,414 & 9,462,022,810 & 1,245 & 105 \\
25 & AIDA* & 4,776,563,020 & 14,967,139,880 & 370 & 105 \\
25 & rAIDA* & 13,621,337,969 & 42,579,832,339 & 791 & 105 \\
\hline
26 & PEM-BAE* & 1,043,423,515 & 3,322,959,911 & 468 & 99 \\
26 & AIDA* & 20,621,200,744 & 66,416,950,731 & 2,212 & 99 \\
26 & rAIDA* & 3,463,364,779 & 10,900,612,650 & 221 & 99 \\
\hline
27 & PEM-BAE* & 352,497,075 & 1,125,759,542 & 165 & 98 \\
27 & AIDA* & 623,798,323 & 2,005,600,613 & 44 & 98 \\
27 & rAIDA* & 3,597,787,376 & 11,382,223,685 & 234 & 98 \\
\hline
28 & PEM-BAE* & 547,971,940 & 1,764,505,585 & 264 & 88 \\
28 & AIDA* & 3,786,487,784 & 12,224,161,369 & 254 & 88 \\
28 & rAIDA* & 8,692,306,706 & 27,879,346,477 & 584 & 88 \\
\hline
29 & PEM-BAE* & 335,757,077 & 1,077,689,262 & 172 & 92 \\
29 & AIDA* & 581,364,286 & 1,854,619,108 & 90 & 92 \\
29 & rAIDA* & 2,696,888,415 & 8,740,333,614 & 328 & 92 \\
\hline
30 & PEM-BAE* & 2,229,244,262 & 7,141,272,421 & 1,014 & 99 \\
30 & AIDA* & 33,327,587,471 & 106,997,263,315 & 2,511 & 99 \\
30 & rAIDA* & 131,207,748,664 & 417,347,549,017 & 9,892 & 99 \\
\hline
31 & PEM-BAE* & 165,410,379 & 534,243,911 & 97 & 97 \\
31 & AIDA* & 943,787,432 & 3,057,123,999 & 81 & 97 \\
31 & rAIDA* & 24,385,419,169 & 79,050,507,238 & 1,682 & 97 \\
\hline
32 & PEM-BAE* & 35,763,541,684 & 115,037,007,664 & 16,898 & 106 \\
32 & AIDA* & 312,137,711,728 & 997,185,135,000 & 29,195 & 106 \\
32 & rAIDA* & 2,491,371,346,301 & 7,990,457,375,484 & 234,616 & 106 \\
\hline
33 & PEM-BAE* & 8,414,844,647 & 26,920,078,806 & 3,792 & 102 \\
33 & AIDA* & 46,451,633,545 & 147,871,894,613 & 3,886 & 102 \\
33 & rAIDA* & 182,236,486,186 & 581,410,425,737 & 14,892 & 102 \\
\hline
34 & PEM-BAE* & 1,741,485,012 & 5,583,334,380 & 760 & 98 \\
34 & AIDA* & 40,024,395,290 & 128,437,310,919 & 4,911 & 98 \\
34 & rAIDA* & 5,277,202,444 & 16,861,802,959 & 325 & 98 \\
\hline
35 & PEM-BAE* & 140,062,316 & 451,723,901 & 84 & 90 \\
35 & AIDA* & 499,903,906 & 1,601,617,848 & 62 & 90 \\
35 & rAIDA* & 341,393,075 & 1,097,793,866 & 23 & 90 \\
\hline
36 & PEM-BAE* & 495,304,641 & 1,594,666,052 & 240 & 100 \\
36 & AIDA* & 2,698,159,455 & 8,677,615,693 & 222 & 100 \\
36 & rAIDA* & 20,062,621,118 & 64,712,694,760 & 1,516 & 100 \\
\hline
37 & PEM-BAE* & 42,448,377 & 133,974,933 & 36 & 96 \\
37 & AIDA* & 176,088,934 & 555,352,440 & 14 & 96 \\
37 & rAIDA* & 215,908,715 & 681,002,670 & 19 & 96 \\
\hline
38 & PEM-BAE* & 14,628,451,144 & 46,816,491,293 & 6,456 & 104 \\
38 & AIDA* & 240,449,464,103 & 771,787,649,108 & 21,090 & 104 \\
38 & rAIDA* & 297,309,206,780 & 952,751,563,739 & 16,566 & 104 \\
\hline
39 & PEM-BAE* & 15,247,477 & 49,575,653 & 21 & 82 \\
39 & AIDA* & 16,659,387 & 54,193,955 & 2 & 82 \\
39 & rAIDA* & 23,765,362 & 77,316,278 & 3 & 82 \\
\hline
40 & PEM-BAE* & 665,359,298 & 2,111,896,270 & 305 & 106 \\
40 & AIDA* & 27,319,752,272 & 87,441,169,355 & 2,350 & 106 \\
40 & rAIDA* & 3,356,759,072 & 10,588,713,901 & 245 & 106 \\
\hline
41 & PEM-BAE* & 9,184,797,027 & 29,192,002,231 & 4,300 & 108 \\
41 & AIDA* & 143,443,190,792 & 459,599,370,997 & 13,669 & 108 \\
41 & rAIDA* & 248,490,162,021 & 787,895,371,136 & 29,490 & 108 \\
\hline
42 & PEM-BAE* & 3,360,156,617 & 10,563,674,884 & 1,446 & 104 \\
42 & AIDA* & 42,701,129,781 & 133,901,689,287 & 3,103 & 104 \\
42 & rAIDA* & 61,675,010,878 & 191,670,808,269 & 5,384 & 104 \\
\hline
43 & PEM-BAE* & 27,655,526 & 86,783,847 & 26 & 93 \\
43 & AIDA* & 424,991,002 & 1,342,993,964 & 87 & 93 \\
43 & rAIDA* & 41,147,319 & 128,315,339 & 6 & 93 \\
\hline
44 & PEM-BAE* & 2,564,119,144 & 8,196,575,888 & 1,117 & 101 \\
44 & AIDA* & 27,292,984,883 & 86,492,573,883 & 3,107 & 101 \\
44 & rAIDA* & 12,585,530,174 & 40,235,026,931 & 757 & 101 \\
\hline
45 & PEM-BAE* & 3,343,626,420 & 10,731,780,830 & 1,479 & 100 \\
45 & AIDA* & 40,044,217,948 & 128,274,186,070 & 8,035 & 100 \\
45 & rAIDA* & 127,779,500,497 & 411,479,697,811 & 10,877 & 100 \\
\hline
46 & PEM-BAE* & 352,424,217 & 1,137,139,590 & 179 & 92 \\
46 & AIDA* & 9,865,063,039 & 31,738,097,602 & 1,534 & 92 \\
46 & rAIDA* & 521,478,728 & 1,689,254,936 & 49 & 92 \\
\hline
47 & PEM-BAE* & 17,429,341,353 & 55,807,506,350 & 8,030 & 107 \\
47 & AIDA* & 283,715,030,330 & 911,965,768,598 & 21,557 & 107 \\
47 & rAIDA* & 243,995,327,026 & 773,713,951,124 & 14,068 & 107 \\
\hline
48 & PEM-BAE* & 2,791,739,131 & 8,937,517,942 & 1,180 & 100 \\
48 & AIDA* & 45,208,243,508 & 144,973,497,292 & 4,089 & 100 \\
48 & rAIDA* & 26,404,070,525 & 84,645,954,669 & 1,670 & 100 \\
\hline
49 & PEM-BAE* & 44,666,558,130 & 141,194,804,106 & 20,222 & 113 \\
49 & AIDA* & 2,255,256,313,124 & 7,127,089,799,036 & 377,304 & 113 \\
49 & rAIDA* & 344,477,416,329 & 1,077,576,140,431 & 25,393 & 113 \\
\hline
\end{tabular}
\end{adjustbox}
\caption{24-STP Instances 20-50}
\label{app:tab:stp_second}
\end{table}

\begin{table}[htb]
\begin{adjustbox}{width=0.93\columnwidth}
\begin{tabular}{|l|l|r|r|r|r|}
\hline
\multicolumn{1}{|c|}{\textbf{Instance}} & \multicolumn{1}{c|}{\textbf{Algorithm}} & \multicolumn{1}{c|}{\textbf{Expanded}} & \multicolumn{1}{c|}{\textbf{Generated}} & \multicolumn{1}{c|}{\textbf{Elapsed}} & \multicolumn{1}{c|}{\textbf{Solution}} \\
\hline
0 & PEM-BAE* & 11,402,376 & 68,414,256 & 167 & 127 \\
0 & PEM-MM & 113,477,535 & 680,865,200 & 369 & 127 \\
0 & PEM-A* & 96,251,817 & 577,510,898 & 111 & 127 \\
0 & PEM-rA* & 66,542,001 & 399,251,994 & 78 & 127 \\
\hline
1 & PEM-BAE* & 25,775,736 & 154,654,416 & 200 & 134 \\
1 & PEM-MM & 351,046,537 & 2,106,279,222 & 567 & 134 \\
1 & PEM-A* & 363,462,637 & 2,180,775,804 & 403 & 134 \\
1 & PEM-rA* & 331,575,951 & 1,989,455,675 & 365 & 134 \\
\hline
2 & PEM-BAE* & 221,402,494 & 1,328,413,066 & 680 & 196 \\
2 & PEM-MM & 3,065,812,530 & 18,394,856,344 & 3,566 & 196 \\
2 & PEM-A* & 2,266,947,910 & 13,601,668,477 & 2,756 & 196 \\
2 & PEM-rA* & 1,886,517,301 & 11,319,100,367 & 2,236 & 196 \\
\hline
3 & PEM-BAE* & 875,748,652 & 5,254,491,321 & 1,407 & 226 \\
3 & PEM-MM & 17,627,193,887 & 105,763,116,151 & 19,716 & 226 \\
3 & PEM-A* & 15,257,726,280 & 91,546,318,447 & 19,595 & 226 \\
3 & PEM-rA* & 18,694,559,963 & 112,167,326,767 & 24,084 & 226 \\
\hline
4 & PEM-BAE* & 427,190,546 & 2,563,142,872 & 908 & 210 \\
4 & PEM-MM & 7,205,812,198 & 43,234,848,104 & 7,584 & 210 \\
4 & PEM-A* & 4,005,698,179 & 24,034,172,187 & 4,766 & 210 \\
4 & PEM-rA* & 5,953,665,000 & 35,721,974,133 & 7,320 & 210 \\
\hline
5 & PEM-BAE* & 146,546,097 & 879,276,416 & 480 & 194 \\
5 & PEM-MM & 2,690,832,340 & 16,144,991,560 & 3,020 & 194 \\
5 & PEM-A* & 1,662,774,594 & 9,976,645,755 & 1,956 & 194 \\
5 & PEM-rA* & 1,720,710,614 & 10,324,262,888 & 1,978 & 194 \\
\hline
6 & PEM-BAE* & 396,001,541 & 2,376,007,755 & 796 & 209 \\
6 & PEM-MM & 6,952,068,090 & 41,712,369,788 & 7,319 & 209 \\
6 & PEM-A* & 5,467,395,121 & 32,804,351,175 & 6,751 & 209 \\
6 & PEM-rA* & 3,525,403,361 & 21,152,383,049 & 4,237 & 209 \\
\hline
7 & PEM-BAE* & 302,449,443 & 1,814,693,233 & 791 & 208 \\
7 & PEM-MM & 5,473,929,538 & 32,843,536,479 & 5,812 & 208 \\
7 & PEM-A* & 3,668,644,117 & 22,011,830,694 & 4,426 & 208 \\
7 & PEM-rA* & 3,530,151,234 & 21,180,889,797 & 4,241 & 208 \\
\hline
8 & PEM-BAE* & 879,254,053 & 5,275,495,463 & 1,486 & 231 \\
8 & PEM-MM & 11,786,153,720 & 70,716,724,653 & 13,013 & 231 \\
8 & PEM-A* & 8,556,073,364 & 51,336,276,445 & 10,663 & 231 \\
8 & PEM-rA* & 10,888,264,080 & 65,329,503,752 & 13,841 & 231 \\
\hline
9 & PEM-BAE* & 37,981,991 & 227,891,945 & 269 & 155 \\
9 & PEM-MM & 481,198,567 & 2,887,191,276 & 802 & 155 \\
9 & PEM-A* & 448,034,637 & 2,688,207,607 & 510 & 155 \\
9 & PEM-rA* & 228,474,350 & 1,370,846,061 & 252 & 155 \\
\hline
10 & PEM-BAE* & 154,672,578 & 928,035,264 & 532 & 184 \\
10 & PEM-MM & 2,283,285,317 & 13,699,688,562 & 2,700 & 184 \\
10 & PEM-A* & 1,896,591,734 & 11,379,529,095 & 2,275 & 184 \\
10 & PEM-rA* & 1,066,926,150 & 6,401,553,672 & 1,224 & 184 \\
\hline
11 & PEM-BAE* & 173,190,498 & 1,039,142,833 & 513 & 190 \\
11 & PEM-MM & 3,476,195,799 & 20,857,156,385 & 3,788 & 190 \\
11 & PEM-A* & 2,013,246,081 & 12,079,473,280 & 2,399 & 190 \\
11 & PEM-rA* & 2,087,343,968 & 12,524,048,476 & 2,480 & 190 \\
\hline
12 & PEM-BAE* & 490,313,689 & 2,941,880,692 & 929 & 209 \\
12 & PEM-MM & 6,302,878,412 & 37,817,233,326 & 6,695 & 209 \\
12 & PEM-A* & 5,279,998,033 & 31,679,962,796 & 6,541 & 209 \\
12 & PEM-rA* & 3,467,306,313 & 20,803,818,254 & 4,139 & 209 \\
\hline
13 & PEM-BAE* & 265,428,652 & 1,592,570,397 & 648 & 190 \\
13 & PEM-MM & 4,039,716,079 & 24,238,274,622 & 4,363 & 190 \\
13 & PEM-A* & 3,480,192,849 & 20,881,150,573 & 4,176 & 190 \\
13 & PEM-rA* & 2,753,420,297 & 16,520,494,513 & 3,287 & 190 \\
\hline
14 & PEM-BAE* & 19,924,463 & 119,546,778 & 155 & 133 \\
14 & PEM-MM & 243,416,450 & 1,460,498,700 & 440 & 133 \\
14 & PEM-A* & 149,535,566 & 897,213,388 & 170 & 133 \\
14 & PEM-rA* & 175,698,215 & 1,054,189,290 & 198 & 133 \\
\hline
15 & PEM-BAE* & 218,669,651 & 1,312,017,745 & 644 & 194 \\
15 & PEM-MM & 3,229,210,445 & 19,375,253,326 & 3,640 & 194 \\
15 & PEM-A* & 1,998,734,210 & 11,992,404,448 & 2,348 & 194 \\
15 & PEM-rA* & 2,039,686,344 & 12,238,108,147 & 2,436 & 194 \\
\hline
16 & PEM-BAE* & 229,045,224 & 1,374,271,197 & 634 & 189 \\
16 & PEM-MM & 3,788,714,202 & 22,732,279,620 & 4,095 & 189 \\
16 & PEM-A* & 3,440,953,837 & 20,645,716,783 & 4,148 & 189 \\
16 & PEM-rA* & 2,821,803,598 & 16,930,818,873 & 3,327 & 189 \\
\hline
17 & PEM-BAE* & 54,740,919 & 328,445,513 & 348 & 177 \\
17 & PEM-MM & 952,740,416 & 5,716,442,305 & 1,374 & 177 \\
17 & PEM-A* & 669,377,980 & 4,016,267,799 & 758 & 177 \\
17 & PEM-rA* & 563,424,177 & 3,380,544,791 & 629 & 177 \\
\hline
18 & PEM-BAE* & 367,828,352 & 2,206,969,889 & 822 & 201 \\
18 & PEM-MM & 7,145,353,473 & 42,872,111,738 & 7,391 & 201 \\
18 & PEM-A* & 7,122,309,993 & 42,733,851,546 & 8,709 & 201 \\
18 & PEM-rA* & 6,231,287,375 & 37,387,710,571 & 7,587 & 201 \\
\hline
19 & PEM-BAE* & 604,182,452 & 3,625,082,948 & 1,065 & 219 \\
19 & PEM-MM & 11,314,109,761 & 67,884,576,473 & 12,139 & 219 \\
19 & PEM-A* & 6,979,257,042 & 41,875,500,224 & 8,643 & 219 \\
19 & PEM-rA* & 9,368,016,548 & 56,208,049,256 & 11,771 & 219 \\
\hline
\end{tabular}
\end{adjustbox}
\caption{TOH Instances}
\label{app:tab:toh}
\end{table}

\end{document}